\documentclass{article}

\usepackage[utf8]{inputenc} 
\usepackage[T1]{fontenc}    
\usepackage{url}            
\usepackage{amsfonts}       
\usepackage{nicefrac}       
\usepackage{microtype}      
\usepackage{graphicx}
\usepackage{subcaption}
\usepackage{booktabs} 
\usepackage{hyperref}
\usepackage{wrapfig}
\usepackage{multirow}
\usepackage{xcolor}
\usepackage{amsmath}
\usepackage{cite}
\usepackage[linesnumbered,ruled,vlined]{algorithm2e}
\usepackage{paralist}

\usepackage{amsthm}
\theoremstyle{definition}
\newtheorem{definition}{Definition}[section]

\newcommand{\scheme}{NetSyn}
\newcommand{\vocab}{\ensuremath{\Sigma_{\mathit{DSL}}}}

\usepackage[accepted]{mlsys2020}

\mlsystitlerunning{Learning Fitness Functions for Machine Programming}

\begin{document}

\twocolumn[
\mlsystitle{Learning Fitness Functions for Machine Programming}



\mlsyssetsymbol{equal}{*}

\begin{mlsysauthorlist}
\mlsysauthor{Shantanu Mandal}{tamu}
\mlsysauthor{Todd Anderson}{intel}
\mlsysauthor{Javier Turek}{intel}
\mlsysauthor{Justin Gottschlich}{intel}
\mlsysauthor{Shengtian Zhou}{intel}
\mlsysauthor{Abdullah Muzahid}{tamu}
\end{mlsysauthorlist}

\mlsysaffiliation{tamu}{Department of Computer Science and Engineering, Texas A\&M University}
\mlsysaffiliation{intel}{Intel Labs}

\mlsyscorrespondingauthor{Abdullah Muzahid}{abdullah.muzahid@tamu.edu}


\vskip 0.3in

\begin{abstract}
%
%

The problem of automatic software generation has been referred to as \emph{machine programming}. In this work, we propose a framework based on genetic algorithms to help make progress in this domain. Although genetic algorithms (GAs) have been successfully used for many problems, one criticism is that hand-crafting GAs \emph{fitness function}, the test that aims to effectively guide its evolution, can be notably challenging. Our framework presents a novel approach to \emph{learn} the fitness function using neural networks to predict values of ideal fitness functions.
We also augment the evolutionary process with a minimally intrusive search heuristic. This heuristic improves the framework's ability to discover correct programs from ones that are approximately correct and does so with negligible computational overhead. We compare our approach with several state-of-the-art program synthesis methods and demonstrate that it finds more correct programs with fewer candidate program generations.
\end{abstract}


]



\printAffiliationsAndNotice{}  
\section{Introduction}
\label{sec-intro}
In recent years, there has been notable progress in the space of automatic software generation, also known as \emph{machine programming} (MP)~\cite{gottschlich:2018:mapl, ratner:2018:sysml}.  An  MP system produces a program as output that satisfies some input specification to the system, often in the form of input-output examples.
The previous approaches to this problem have ranged from formal program synthesis~\cite{gulwani12, alur15} to machine learning (ML)~\cite{Balog:2017:github, Zohar:2018:nips, robustfill, npi} as well as their combinations~\cite{feng18}.  Genetic algorithms (GAs) have also been shown 
to have significant promise for MP~\cite{aip, stackgp, lgp, allgp}. GA is a simple and intuitive approach and demonstrates competitive performance in many challenging domains~\cite{Korns2011, such17, Read:2018:dblp}. Therefore, in this paper, we focus on GA - more specifically, a fundamental aspect of GA in the context of MP.

A \emph{genetic algorithm} (GA) is a machine learning technique that attempts to solve a problem from a pool of candidate solutions. These generated candidates are iteratively evolved and mutated and selected for survival based on a grading criteria, called the \emph{fitness function}. Fitness functions are usually hand-crafted heuristics that grade the approximate correctness of candidate solutions such that those that are closer to being correct are more likely to appear in subsequent generations.

In the context of MP, candidate solutions are programs, initially random but evolving over time to get closer to a program satisfying the input specification.  Yet, to guide that evolution, it is particularly difficult to design an effective fitness function for a GA-based MP system.  The fitness function is given a candidate program and the input specification (e.g., input-output examples) and from those, must estimate how close that candidate program is to satisfying the specification.  However, we know that a program having only a single mistake may produce output that in no \emph{obvious} way resembles the correct output.  
That is why, one of the most frequently used fitness functions (i.e., edit-distance between outputs) in this domain~\cite{aip, stackgp, lgp, allgp} will in many cases give wildly wrong estimates of candidate program correctness.
Thus, it is clear that designing effective fitness functions for MP are difficult.

Designing simple and effective fitness functions is a unique challenge for GA.
Despite many successful applications of GA, it still remains an open challenge to automate
the generation of such fitness functions.
An impediment to this goal is that fitness function complexity tends to increase proportionally with the problem being solved, with MP being particularly complex. In this paper, we explore an approach to automatically generate these fitness functions by representing their structure with a neural network.  While we investigate this technique
in the context of MP, 
we believe the technique to be applicable and generalizable to other domains.  We make 
the following technical contributions:
\begin{itemize}
    \item {\em Fitness Function:} Our fundamental contribution is in the automation of fitness functions for genetic algorithms. We propose to do so by mapping fitness function generation as a big data learning problem. To the best of our knowledge, our work is the {\em first} of its kind to use a neural network as a genetic algorithm's fitness function for the purpose of MP. 
    
    \item {\em Convergence:} A secondary contribution is in our utilization of local neighborhood search to improve the convergence of approximately correct candidate solutions. We demonstrate its efficacy empirically.
    
    \item {\em Generality:} We demonstrate that our approach can support different neural network fitness functions, uniformly. We develop a neural network model to predict the fitness score based on the given specification and program trace.
    
    \item {\em Metric:} We contribute a new metric suitable for MP domain. The metric, ``search space'' size (i.e., how many candidate programs have been searched), is an alternative to program generation time, and is designed to emphasize the algorithmic efficiency as opposed to the implementation efficiency of an MP approach.
    
\end{itemize}

\section{Related Work}
\label{sec-rel-work}

Machine programming can be achieved in many ways. One way is by using \emph{formal program synthesis}, a technique that uses formal methods and rules to generate programs~\cite{mana75}. Formal program synthesis usually guarantees some program properties by evaluating a generated program's semantics against a corresponding specification~\cite{gulwani12, alur15}. Although useful, such formal synthesis techniques can often be limited by exponentially increasing computational overhead that grows 
with the program's instruction size~\cite{Heule:2016:pldi, Bodik:2013:sttt, Solar-Lezama:2006:asplos, Loncaric:2018:icse, Cheung:2012:cikm}.

An alternative to formal methods for MP is to use machine learning (ML). Machine learning differs from traditional formal program synthesis in that it generally does not provide correctness guarantees. Instead, ML-driven MP approaches are usually only \emph{probabilistically} correct, i.e., their results are derived from sample data relying on statistical significance~\cite{Murphy:2012:mitpress}. Such ML approaches tend to explore software program generation using an objective function. Objective functions are used to guide an ML system's exploration of a problem space to find a solution. 

More recently, there has been a surge of research exploring ML-based MP using neural networks (NNs). For example, in \cite{deepcoder}, the authors train a neural network with input-output examples to predict the probabilities of the functions that are most likely to be used in a program. Raychev et al.~\cite{raychev14} take a different approach and use an n-gram model 
to predict the functions that are most likely to complete a partially constructed program.
Robustfill~\cite{robustfill} encodes input-output examples using a series of 
recurrent neural networks (RNN), and generates the the program using another RNN one token at a time.
Bunel et al.~\cite{bunel18} explore a unique approach that combines reinforcement learning (RL) with a supervised model to find semantically correct programs. These are only a few of the works in the MP space using neural networks~\cite{npi, cai17, Chen18}.

Significant research has been done in the field of genetic programming~\cite{stackgp, lgp, allgp} whose goal is to find a solution in the form of a complete or partial program for a given specification.  Prior work in this field has tended to focus on either the representation of programs or operators during the evolution process. 
Real et al.~\cite{real18} recently demonstrated that genetic algorithms can generate accurate image classifiers. Their approach produced a state-of-the-art classifier for CIFAR-10~\cite{cifar10} and ImageNet~\cite{imagenet} datasets. Moreover, genetic algorithms have been exploited to successfully automate the neural architecture optimization process~\cite{salimans17, such17, liu17, santientlabs, real2020automlzero}. Even with this notable progress, genetic algorithms can be challenging to use due to the complexity of hand-crafting fitness functions that guide the search. We claim that our proposed approach is the first of its kind to automate the generation of fitness functions.

\begin{figure*}[htpb]
    \begin{center}
        \includegraphics[width=0.8\textwidth]{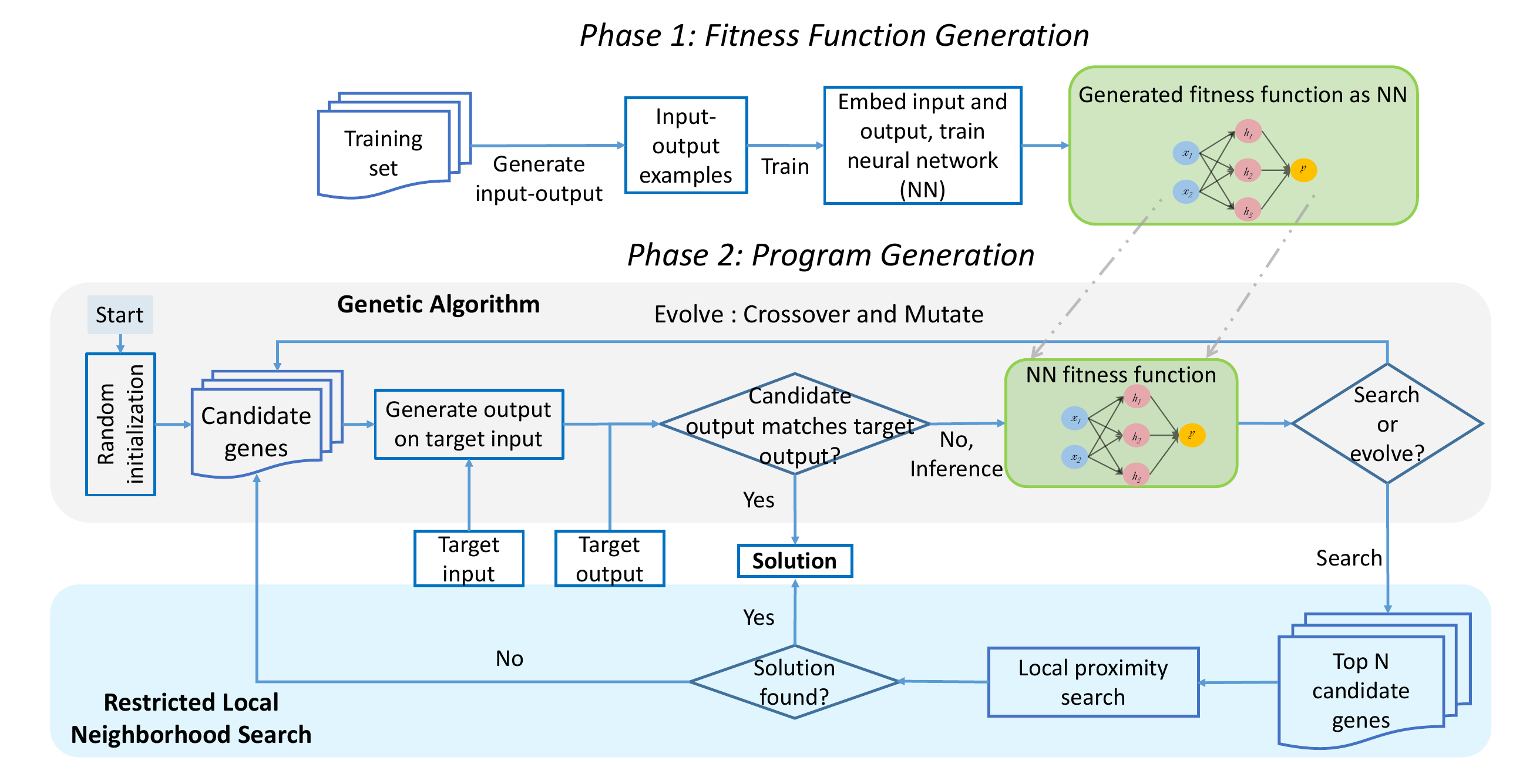}
        \caption{Overview of \scheme. Phase 1 automates the fitness function generation by training a neural network on a corpus of example programs and their inputs and outputs. Phase 2 finds the target program for a given input-output example using the trained neural network as a fitness function in a genetic algorithm.}
        \label{fig-overview}
    \end{center}
\end{figure*}

\section{Background}
\label{sec-problem}
Let $S^{t} = \{(I_j, O^t_j)\}^m_{j=1}$ be a set of $m$ input-output pairs, such that the output $O_j^t$ is obtained by executing the program $P^t$ on the input $I_j$. 
Inherently, the set $S^{t}$ of input-output examples describes the behavior of the program $P^t$.
One would like to synthesize a program $P^{t'}$ that recovers the same functionality of $P^t$. However, $P^t$ is usually unknown, and we are left with the set $S^{t}$, which was obtained by running $P^t$.
Based on this assumption, we define equivalency between two programs as follows:
\begin{definition}[Program Equivalency]
\label{def:equiv}
Programs $P^a$ and $P^b$ are equivalent under the set $S = \{(I_j, O_j)\}^m_{j=1}$ of input-output examples if and only if 
$P^a(I_j) = P^b(I_j) = O_j$, for $1 \le j \le m$. We denote the equivalency by $P^a \equiv_{S} P^b$.
\end{definition}
Definition \ref{def:equiv} suggests that to obtain a program equivalent to $P^t$, we need to synthesize a program that is consistent with the set $S^{t}$.
Therefore, our goal is find a program $P^{t'}$ that is equivalent to the target program $P^t$ (which was used to generate $S^t$), i.e., $P^{t'} \equiv_{S^{t}} P^t$.
This task is known as Inductive Program Synthesis (IPS).
As suggested by \cite{deepcoder}, a machine learning based solution to the IPS problem requires the definition of some components. First, we need a programming language that defines the domain of valid programs. 
Second, we need a method to search over the program domain. The search method sweeps over the program domain to find $P^{t'}$ that satisfies the equivalency property. Optionally, we may want to define a ranking function to rank all the  solutions found and choose the best ones. Last, as we plan to base our solution on machine learning techniques, we will need data
to train  models.

\section{\scheme}
\label{sec-main}

Here, we describe our solution to IPS  in more detail, including the choices and novelties for each of the proposed components. We name our solution \scheme{} as it is based on neural networks for program synthesis. 

\subsection{Domain Specific Language}
\label{sec-domain}
As \scheme{}'s programming language, we choose a domain specific language (DSL) constructed specifically for it. This choice allows us to constrain the program space by restricting the operations used by our solution.
\scheme's DSL follows the DeepCoder's DSL \cite{deepcoder}, which was inspired by SQL and LINQ~\cite{Kulkarni:2007}. 
The only data types in the language are \emph{(i)} integers and  \emph{(ii)} lists of integers.
The DSL contains 41 functions, each taking one or two arguments and returning one output. Many of these functions include operations for list manipulation. Likewise, some operations also require lambda functions. There is no explicit control flow (conditionals or looping) in the DSL. However, several of the operations are high-level functions and are implemented using such control flow structures. A full description of the DSL can be found in the supplementary material. 
With these data types and operations, we define a program $P$ as a sequence of functions. Table \ref{table:example} presents an example of a program of 4 instructions with an input and respective output.


Arguments to functions are not specified via named variables. Instead, each function uses the output of the previously executed function that  produces the type of output that is used as the input to the next function. The first function of each program uses the provided input $I$. If $I$ has a type mismatch, default values are used (i.e., 0 for integers and an empty list for a list of integers). The final output of a programs is the output of its last function. 

\begin{table}[htpb]
  \caption{An example program of length 4 with an input and corresponding output.}
  \label{table:example}
   \centering
    \scalebox{1}{
  \begin{tabular}{ll}
  \hline
   [int] & Input: \\
   \textsc{Filter ($>$0)}  & [-2, 10, 3, -4, 5, 2] \\
   \textsc{Map (*2)}  &  \\
   \textsc{Sort}  & Output:\\
   \textsc{Reverse}  & [20, 10, 6, 4] \\\hline

  \end{tabular} }
\end{table}

As a whole, \scheme's DSL is novel and amenable to genetic algorithms. The language is defined such that all possible programs are \emph{valid by construction}. This makes the whole program space valid and is important to facilitate the search of programs by any learning method. 
In particular, this is very useful in evolutionary process in genetic algorithms. When genetic crossover occurs between two programs or mutation occurs within a single program, the resulting program will \emph{always} be valid. This eliminates the need for pruning to identify valid programs.




\subsection{Search Process}
\scheme\ synthesizes a program by searching the program space with a genetic algorithm-based method~\cite{globalopt}. It does this by creating a population of random genes (i.e., candidate programs) of a given length $L$ and uses a learned neural network-based fitness function (NN-FF) to estimate the fitness of each gene. Higher graded genes are preferentially selected for crossover and mutation to produce the next generation of genes. In general, \scheme\ uses this process to evolve the genes from one generation to the next until it discovers a correct candidate program as verified by the input-output examples. From time to time, \scheme\ takes the top $N$ scoring genes from the population, determines their neighborhoods, and looks for the target program using a local proximity search. If a correctly generated program is not found within the neighborhoods, the evolutionary process resumes. Figure~\ref{fig-overview} summarizes the \scheme{}'s search process. 

We use a value encoding approach for each gene. A gene $\zeta$ is represented as a sequence of values from \vocab{}, the set of functions. Formally, a gene $\zeta = \left(f_1,\dots,f_i,\dots,f_L \right)$, where $f_i \in \vocab{}$. Practically, each $f_i$ contains an identifier (or index) corresponding to one of the DSL functions. The encoding scheme satisfies a one-to-one match between programs and genes. 

The search process begins with a set $\Phi^0$ of $|\Phi^0|=T$ randomly generated programs.
If a program equivalent to the target program $P^t$ is found, the search process stops.
Otherwise, the genes are ranked using a learned NN-FF. A small percentage (e.g., 20\%) of the top graded genes in $\Phi^j$ are passed in an unmodified fashion to the next generation $\Phi^{j+1}$ for the next evolutionary phase. This guarantees that some of the top graded genes are identically preserved, aiding in forward progress guarantees. The remaining genes of the new generation $\Phi^{j+1}$ are created through crossover or mutation with some probability. For crossover, two genes from $\Phi^j$ are selected using the Roulette Wheel algorithm with the crossover point selected randomly~\cite{Goldberg:1989:awlp}. 
For mutation, one gene is Roulette Wheel selected and the mutation point $k$ in that gene is selected based on the same learned NN-FF.
The selected value $z_k$ is mutated to some other random value $z'$ such that $z' \in \vocab{}$ and $z'\ne z_k$. 

Crossovers and mutations can occasionally lead to a new gene with dead code. To address this issue, we eliminate dead code. Dead code elimination (DCE)
is a classic compiler technique to remove code from a program that has no effect on the program's output~\cite{Debray:2000:acm}. Dead code is possible in our list DSL if the output of a statement is never used. We implemented DCE in \scheme\ by tracking the input/output dependencies between statements and eliminating those statements whose outputs are never used.  \scheme\ uses DCE during candidate program generation and during crossover/mutation to ensure that the effective length of the program is not less than the target program length due to the presence of dead code.
If dead code is present, we repeat crossover and mutation until a gene without dead code is produced. 



\begin{figure*}[htpb]
    \centering
    
    \begin{subfigure}{0.75\textwidth}
        \includegraphics[width=\textwidth]{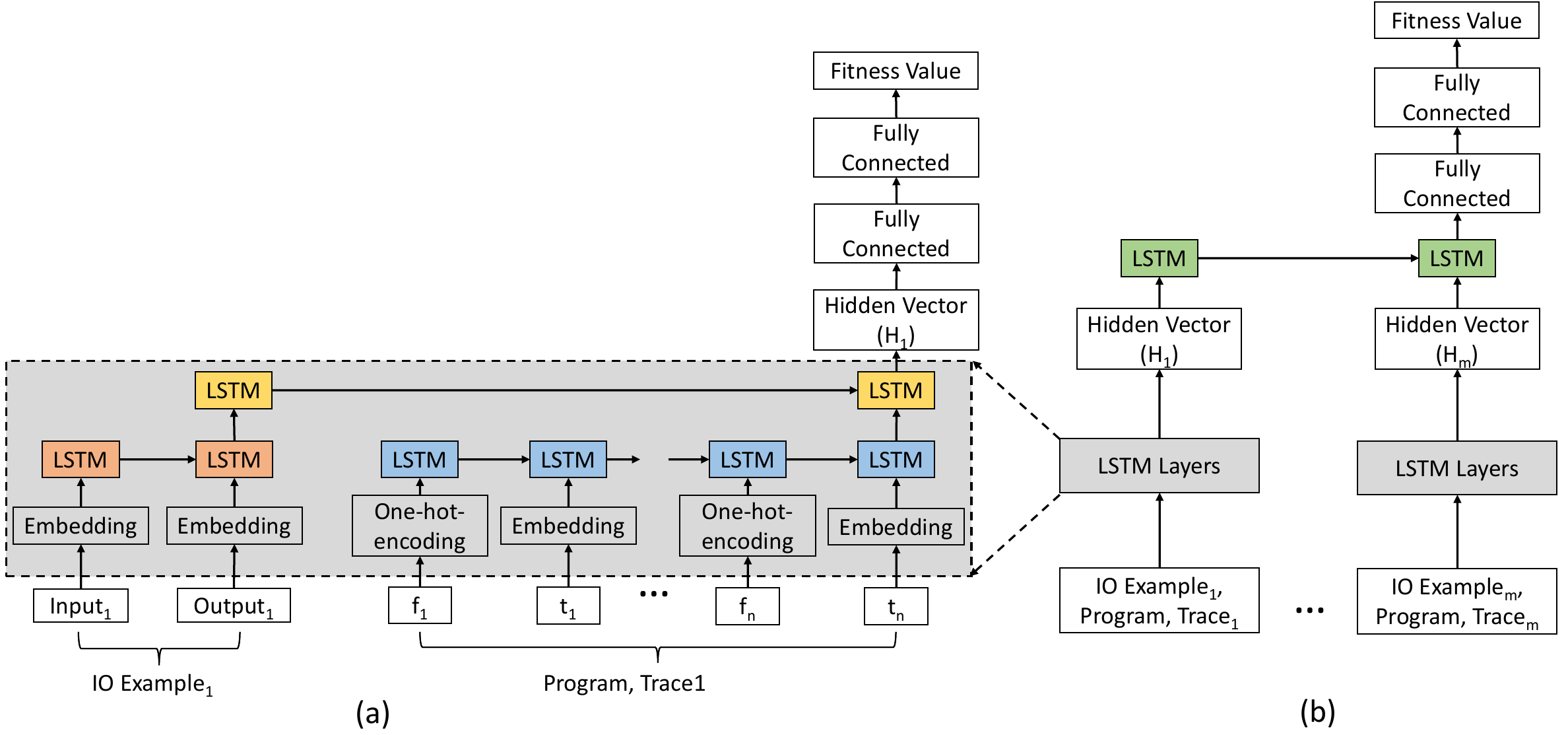}
    \end{subfigure}

    \caption{Neural network fitness function for (a) single and (b) multiple IO examples. In each figure, layers of LSTM encoders are used to combine multiple inputs into hidden vectors for the next layer. Final fitness score is produced by the fully connected layer.}
    \label{fig:model}
\end{figure*}

\subsubsection{Learning the Fitness Function}
\label{sec-fitness}

Evolving the population of genes in a genetic algorithm requires a fitness function to rank the fitness (quality) of genes based on the problem being solved. Ideally, a fitness function should measure how close a gene is to the solution. Namely, it should measure how close a candidate program is to an equivalent of $P^t$ under $S^t$. Finding a good fitness function is of great importance to reduce the
number of steps in reaching the solution and directing the algorithm in the right direction so that 
genetic algorithm are more likely to find $P^t$. 

{\bf Intuition:} A fitness function, often, is handcrafted to approximate some ideal function that is impossible (due to incomplete knowledge about the solution) or too computationally intensive to implement in practice. For example, if we knew $P^t$ beforehand, we could have designed an ideal fitness function that compares a candidate program with $P^t$ and calculates some metric of closeness (e.g., edit distance, the number of common functions etc.) as the fitness score. Since we do not know $P^t$, we cannot implement the ideal fitness function.
Instead, in this work, we propose to approximate the ideal fitness function by learning it from training data (generated from a number of known programs).
For this purpose, we use a neural network model. We train it with the goal of predicting the values of an ideal fitness function.  We call such an ideal fitness function (that would always give the correct answer with respect to the actual solution) the \emph{oracle} fitness function as it is impossible to achieve in practice merely by examining input-output examples.
In this case, our models will not be able to approach the 100\% accuracy of the \emph{oracle} but rather will still have sufficiently high enough accuracy to allow the genetic algorithm to make forward progress.
Also, we note that the trained model needs to generalize to predict for any unavailable solution and not a single specific target case.

We follow ideas from works that have explored the automation of fitness functions using neural networks for approximating a known mathematical model. For example, Matos Dias et al.~\cite{Dias:2014:cejor} automated them for IMRT beam angle optimization, while Khuntia et al.~\cite{Bonomali:2005:motl} used them for rectangular microstrip antenna design automation. In contrast, our work is fundamentally different in that we use a large corpus of program metadata to train our models to predict 
how close a given, incorrect solution could be from an {\em unknown} correct solution (that will generate the correct output).
In other words, we propose to automate the generation of fitness functions using big data learning.
To the best of our knowledge, \scheme\ is the {\em first} proposal for automation of fitness functions in genetic algorithms. In this paper, we demonstrate this idea using MP as the use case.

Given the input-output samples $S^t=\left\{\left(I_j, O_j^t\right)\right\}_j$ of the target program $P^t$ and an ideal fitness function $fit(\cdot)$, we would like a model that predicts the fitness value $fit(\zeta, P^t)$ for a gene $\zeta$. In practice, our model predicts the values of $fit(\cdot)$ from input-output samples in $S^t$ and from execution traces of the program $P^\zeta$ (corresponding to $\zeta$) by running with those inputs. Intuitively, execution traces provide insights of whether the program $P^\zeta$ is on the right track.

In \scheme{}, we use a neural network to model the fitness function, referred to as NN-FF. This task requires us to  generate a training dataset of programs with respective input-output samples. To train the NN-FF, we randomly generate a set of example programs, $E=\left\{P^{e_j}\right\}$, along with a set of random inputs $I^j=\{I_i^{e_j}\}$ per program $P^{e_j}$. We then execute each program $P^{e_j}$ in $E$ with its corresponding input set $I^j$ to calculate the output set $O^j$. Additionally, for each $P^{e_j}$ in $E$, we randomly generate another program $P^{r_j}=(f^{r_j}_{1}, f^{r_j}_{2}, ..., f^{r_j}_{n})$, where $f^{r_j}_{k}$ is a function from the DSL i.e., $f^{r_j}_{k} \in~\vocab{}$. 
We apply the previously generated input $I_i^{e_j}$ to $P^r_j$ to get an execution trace, $T^{rj}_i=(t^{rj}_{i1}, t^{rj}_{i2}, ..., t^{rj}_{in})$, where $t^{rj}_{ik}=f^{rj}_{k}(t^{rj}_{i(k-1)})$ with $t^{rj}_{i1}=f^{rj}_{1}(I_i^{e_j})$ and $t^{rj}_{in}=f^{rj}_{n}(t^{rj}_{i(n-1)})=P^{r_j}(I_i^{e_j})$. Thus, the input set $I^j=\{I^{e_j}_i\}$ of the program $P^{e_j}$ produces a set of traces $T^j=\{T^{r_j}_i\}$ from the program $P^{r_j}$.
We then compare the programs $P^{r_j}$ and $P^{e_j}$ 
to calculate the fitness value and use it as an example to train the neural network. 

In \scheme, the inputs of NN-FF consist of input-output examples, generated programs, and their execution traces. Let us consider the case of a single input-output example, $(I^{e_j}_i, O^{e_j}_i)$. Let us assume that $P^{e_j}$ is the target program that \scheme\ attempts to generate and in the process, it generates $P^{r_j}$ as a potential equivalent. NN-FF uses $(I^{e_j}_i$, $O^{e_j}_i)$, and $\{(f^{r_j}_{k}, t^{r_j}_{ik})\}$ as the inputs. Each of $(I^{e_j}_i$, $O^{e_j}_i)$, and $t^{r_j}_{ik}$ are passed through an embedding layer followed by an LSTM encoder. $f^{r_j}_{k}$ is passed as a one-hot-encoding vector. Figure~\ref{fig:model}(a) shows the NN-FF architecture for a single input-output example. Two layers of LSTM encoders combines the vectors to produce a single vector, $H^j_i$, which is then processed through fully connected layers to predict the fitness value. In order to handle a set of input-output examples, $\{(I^{e_j}_i, O^{e_j}_i)\}$, a set of execution traces, $T^j=\{T^{r_j}_i\}$, is collected from a single generated program, $P^{r_j}$. Each input-output example along with the corresponding execution trace produces a single vector, $H^j_i$. An LSTM encoder combines such vectors to produce a single vector, which is then processed by fully connected layers to predict the fitness value (Figure~\ref{fig:model}(b)).

{\bf Example:} To illustrate, suppose the program in Table~\ref{table:example} is in $E$. Let us assume that $P^{r_j}$ is another program \{\textsc{[int], Filter ($>$0), Map (*2), Reverse, Drop (2)}\}. If we use the input in Table~\ref{table:example} (i.e., [-2, 10, 3, -4, 5, 2]) with $P^{r_j}$, the execution trace is \{{[10, 3, 5, 2], [20, 6, 10, 4], [4, 10, 6, 20], [6, 20]\}. So, the input of NN-FF is \{[-2, 10, 3, -4, 5, 2], [20, 10, 6, 4], $Filter_{v}$, [10, 3, 5, 2], $Map_{v}$, [20, 6, 10, 4], $Reverse_v$, [4, 10, 6, 20], $Drop_v$, [6, 20]\}. $f_v$ indicates the value corresponding to the function $f$.


There are different ways to quantify how close two programs are to one another.  Each of these different methods then has an associated metric and ideal fitness value.  We investigated three such metrics -- common functions, longest common subsequence, and function probability -- which we use as the expected predicted output for the NN-FF.

{\bf Common Functions: }
\scheme\ can use the number of common functions (CF) between $P^\zeta$ and $P^t$ as a fitness value for $\zeta$. In other words, the fitness value of $\zeta$ is
$f^{CF}_{P^t}(\zeta)=|\mathbf{elems}(P^\zeta) \cap \mathbf{elems}(P^t)|$. For the earlier example, $f^{CF}$ will be 3.
Since the output of the neural network will be an integer from 0 to $\mathbf{len}(P_t)$, the neural network can be designed as a multiclass classifier with a softmax layer as the final layer.

{\bf Longest Common Subsequence: }
As an alternative to CF, we can use longest common subsequence (LCS) between $P^\zeta$ and $P^t$. The fitness score of $\zeta$ is 
$f^{LCS}_{P^t}(\zeta)=\mathbf{len}(\mathit{LCS}(P^\zeta, P^t))$.
Similar to CF, training data can be constructed from $E$ which is then fed into a neural network-based multiclass classifier. For the earlier example, $f^{LCS}$ will be 2.

{\bf Function Probability: }
The work \cite{deepcoder} proposed a probability map for the functions in the DSL. 
Let us assume that the probability map $\mathbf{p}$ is defined as the probability of each DSL operation to be in $P^t$ given the input-output samples. Namely, $\mathbf{p} = (p_1, \dots, p_k,\dots,p_{|\vocab{}|})$ such that $p_k=Prob(\mathrm{op}_k \in \mathbf{elems}(P^t)|\{(I_j, O^t_j)\}^m_{j=1})$, where $\mathrm{op}_k$ is the $k^{th}$ operation in the DSL. 
Then, a multiclass, multilabel neural network classifier with sigmoid activation functions used in the output of the last layer can be used to predict the probability map. Training data can be constructed for the neural network using $E$. We can use the probability map to calculate the fitness score of $\zeta$ as 
$f^{FP}_{P^t}(\zeta)=\sum_{k:\mathrm{op}_k \in \mathbf{elems}(P^\zeta)} p_k$.
\scheme\ also uses the probability map to guide the mutation process. For example, instead of mutating a function $z_k$ with $z'$ that is selected randomly, \scheme\ can select $z'$ using Roulette Wheel algorithm using the probability map.


\subsubsection{Local Neighborhood Search}
\label{sec-neighbor}


Neighborhood search (NS) checks some candidate genes in the {\em neighborhood} of the $N$ top scoring genes from the genetic algorithm. The intuition behind NS is that if the target program $P^t$ is in that neighborhood, \scheme\ may be able to find it without relying on the genetic algorithm, which would likely result in a faster synthesis time.

\begin{figure}[htpb]
    \centering
    
    \begin{subfigure}{0.25\textwidth}
        \includegraphics[width=\textwidth]{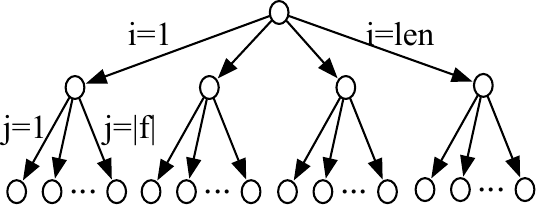}
        \caption{BFS-based}
    \end{subfigure}
    \begin{subfigure}{0.25\textwidth}
        \includegraphics[width=0.95\textwidth]{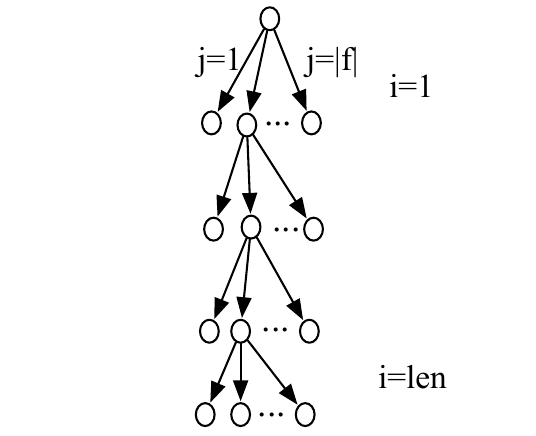}
        \caption{DFS-based}
    \end{subfigure}

    \caption{Examples of neighborhood using (a) BFS- and (b) DFS-based approach. Each neighborhood constructs a set of close-by genes by systematically changing one function at a time.}
    \label{fig:neighbor}
\end{figure}
%
Let us assume that \scheme{} has completed $l$ generations. Then, let $\mu_{l-w+1,l}$ denote the average fitness score of genes for the last $w$ generations (i.e., from $l-w+1$ to $l$) and $\mu_{1,l-w}$ will denote the average fitness score before the last $w$ generations (i.e., from $1$ to $l-w$). Here, $w$ is the sliding window. \scheme\ invokes NS if $\mu_{l-w+1,l} \le \mu_{1,l-w}$. The rationale is that under these conditions, the search procedure has not produced improved genes for the last $w$ generations (i.e., saturating). Therefore, it should check if the neighborhood contains any program equivalent to $P^t$.

\begin{algorithm}[htb]
\footnotesize
\KwIn{A set $G$ of top $N$ scoring genes}
\KwOut{$P^{t'}$, if found, or \emph{Not found} otherwise}
\For{Each $\zeta \in G$}{
  $NH \gets \emptyset$
  
  \For{$i \gets 1~\mathbf{to~len}(\zeta)$}{
    \For{$j \gets 1~\mathbf{to} |\Sigma_{DSL}|$}{
      $\zeta_n \gets \zeta~with~\zeta_i~replaced~with~\mathrm{op}_j$
      $~~such~that~\zeta_i \ne \mathrm{op}_j$
      
      $NH \gets NH \cup \{ \zeta_n\}$
    }
  }
  \If{there is $P^{t'}\in NH$ such that $P^{t'}\equiv_{S^t} P^t$}{
  \Return{$P^{t'}$}
  }
}
\Return{Not found}
\caption{Defines and searches neighborhood based on BFS principle}
\label{algo-neighbor}
\end{algorithm}

{\bf Neighborhood Definition: }
Algorithm~\ref{algo-neighbor} shows how to define and search a neighborhood. The algorithm is inspired by the breadth first search (BFS) method. For each top scoring gene $\zeta$, \scheme\ considers one function at a time starting from the first operation of the gene to the last one. Each selected operation is replaced with all other operations from $\vocab{}$, and inserts the resultant genes into the neighborhood set $NH$. If a program $P^{t'}$ equivalent to $P^t$ is found in $NH$, \scheme\ stops there and returns the solution. Otherwise, it continues the search and returns to the genetic algorithm. The complexity of the search is $\mathcal{O}(N\cdot \mathbf{len}(\zeta)\cdot|\vocab{}|)$, which is significantly smaller than the exponential search space used by a traditional BFS algorithm.
Similar to BFS, \scheme\ can define and search the neighborhood using an approach similar to depth first search (DFS). It is similar to Algorithm~\ref{algo-neighbor} except $i$ keeps track of depth here. After the loop in line 4 finishes, \scheme\ picks the best scoring gene from $NH$ to replace $\zeta$ before going to the next level of depth. The algorithmic complexity remains the same. Figure~\ref{fig:neighbor}(a) and (b) show examples of neighborhood using BFS- and DFS-based approach.

\begin{figure*}[!ht]
    \centering
    \begin{subfigure}{0.3\textwidth}
        \includegraphics[width=\textwidth]{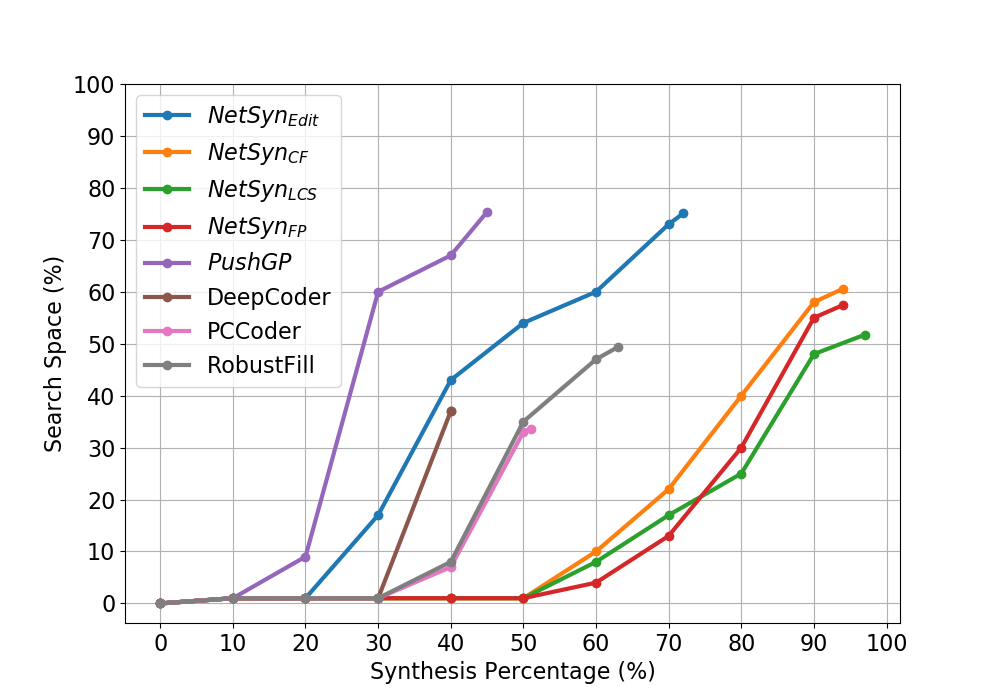}
    \vspace{-0.6cm}
    \caption{Program length = 5}
    
    \end{subfigure}
    ~
    \begin{subfigure}{0.3\textwidth}
        \includegraphics[width=\textwidth]{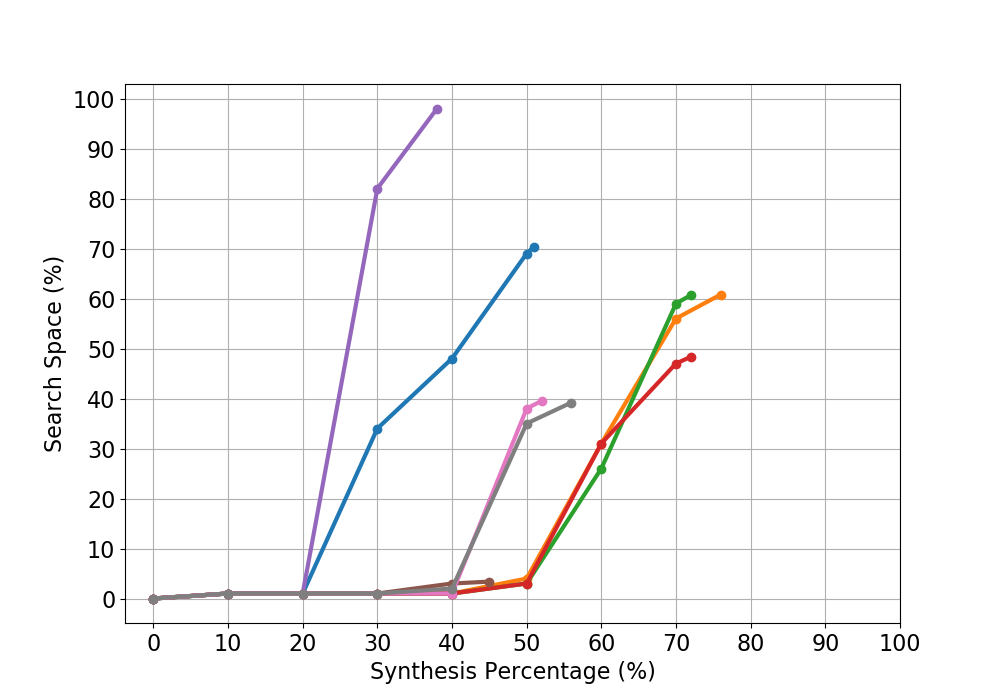}
     \vspace{-0.6cm}
     \caption{Program length = 7}
    \end{subfigure}
    ~
    \begin{subfigure}{0.3\textwidth}
        \includegraphics[width=\textwidth]{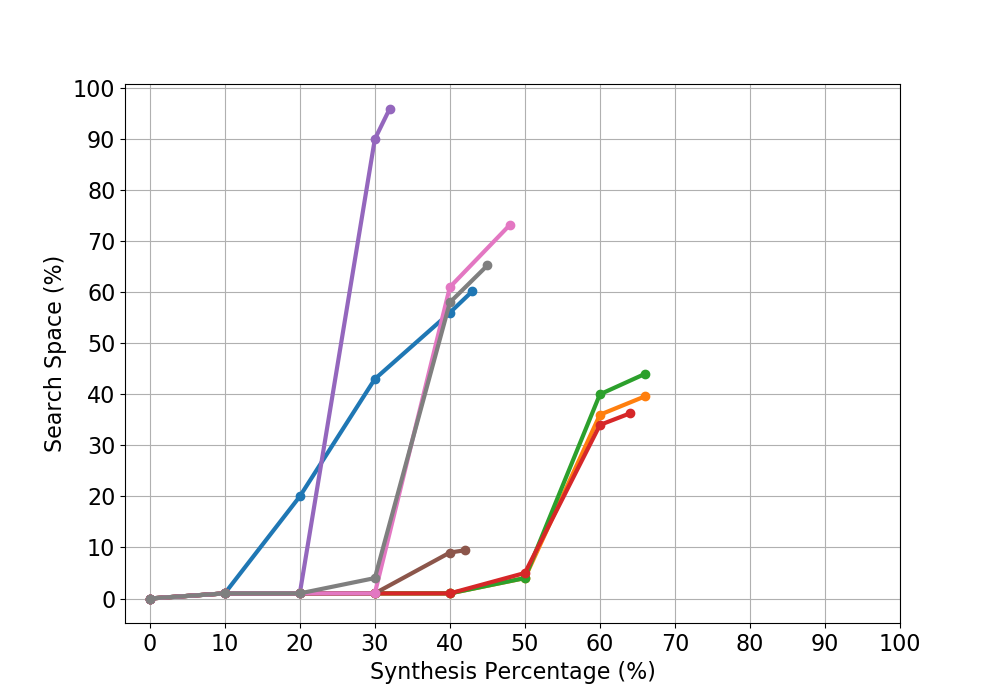}
     \vspace{-0.6cm}
     \caption{Program length = 10}
    \end{subfigure}
    \begin{subfigure}{0.3\textwidth}
        \includegraphics[width=\textwidth]{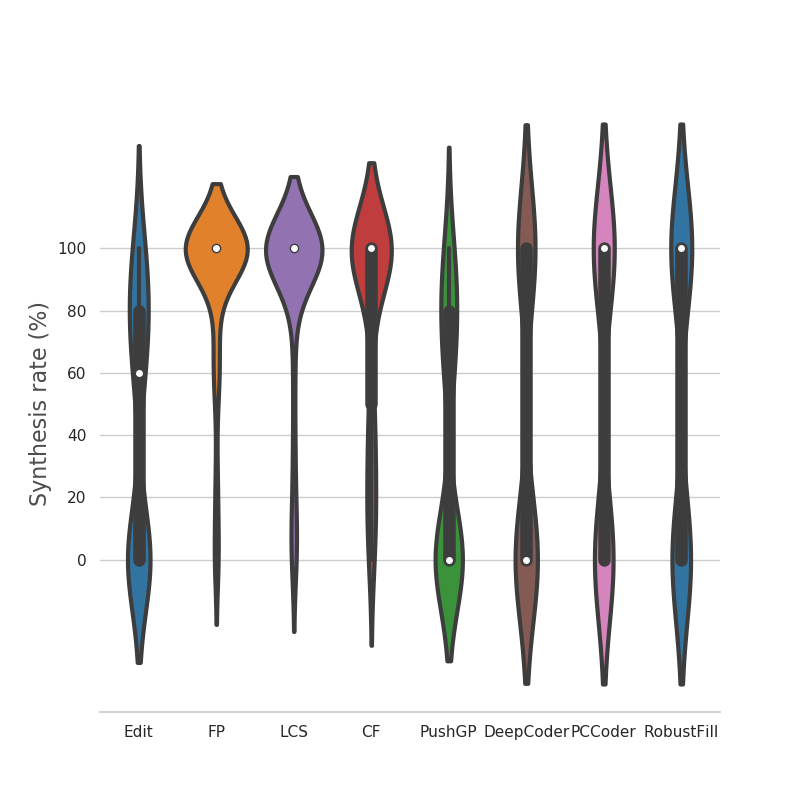}
    \vspace{-0.6cm}
    \caption{Program length = 5}
    \end{subfigure}
    ~
    \begin{subfigure}{0.3\textwidth}
        \includegraphics[width=\textwidth]{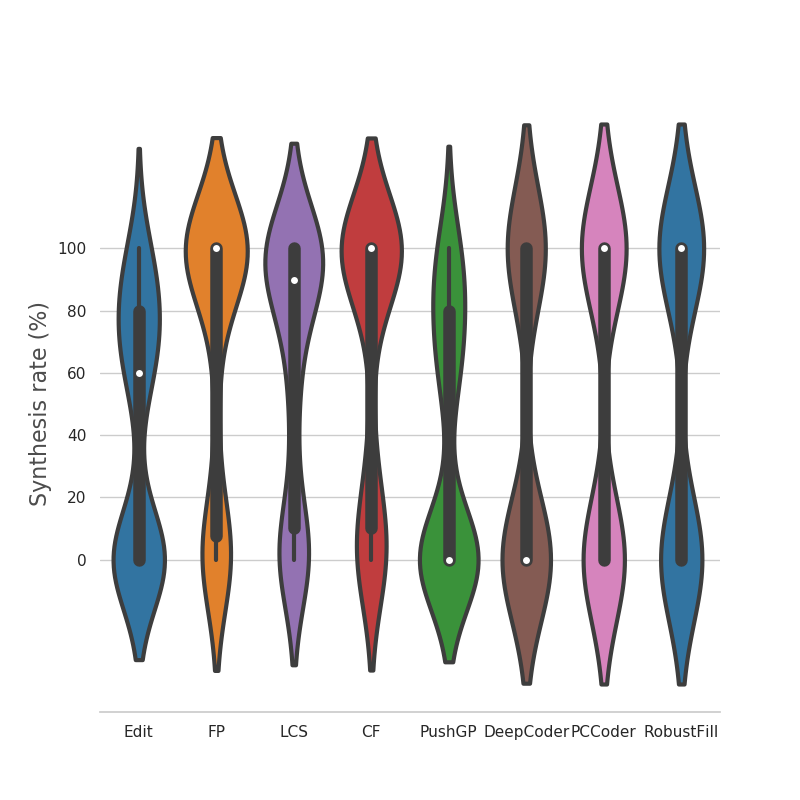}
     \vspace{-0.6cm}
     \caption{Program length = 7}
    \end{subfigure}
    ~
    \begin{subfigure}{0.3\textwidth}
        \includegraphics[width=\textwidth]{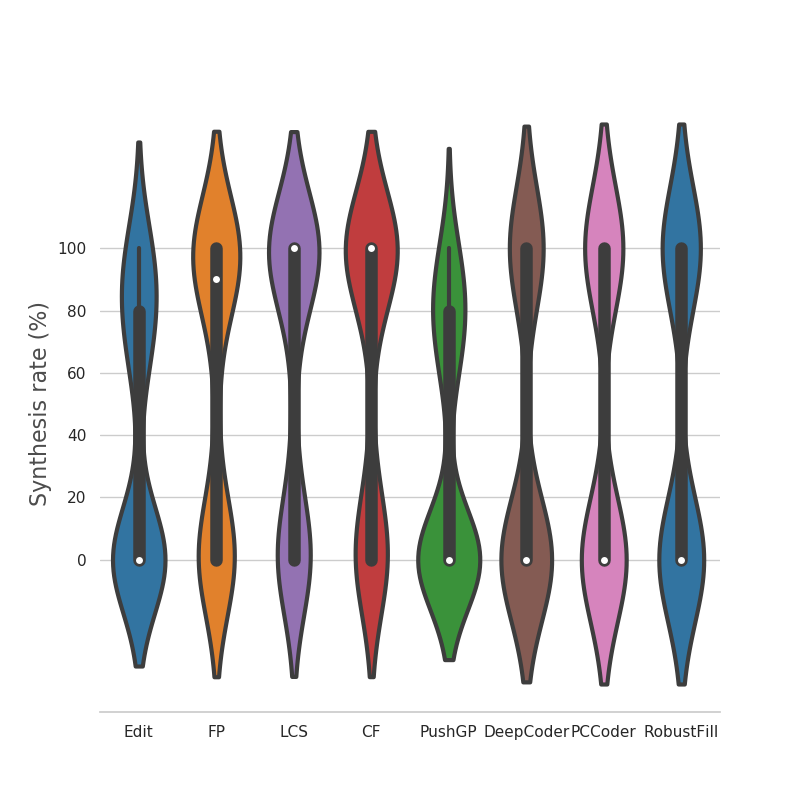}
     \vspace{-0.6cm}
     \caption{Program length = 10}
    \end{subfigure}    
    \begin{subfigure}{0.3\textwidth}
        \includegraphics[width=\textwidth]{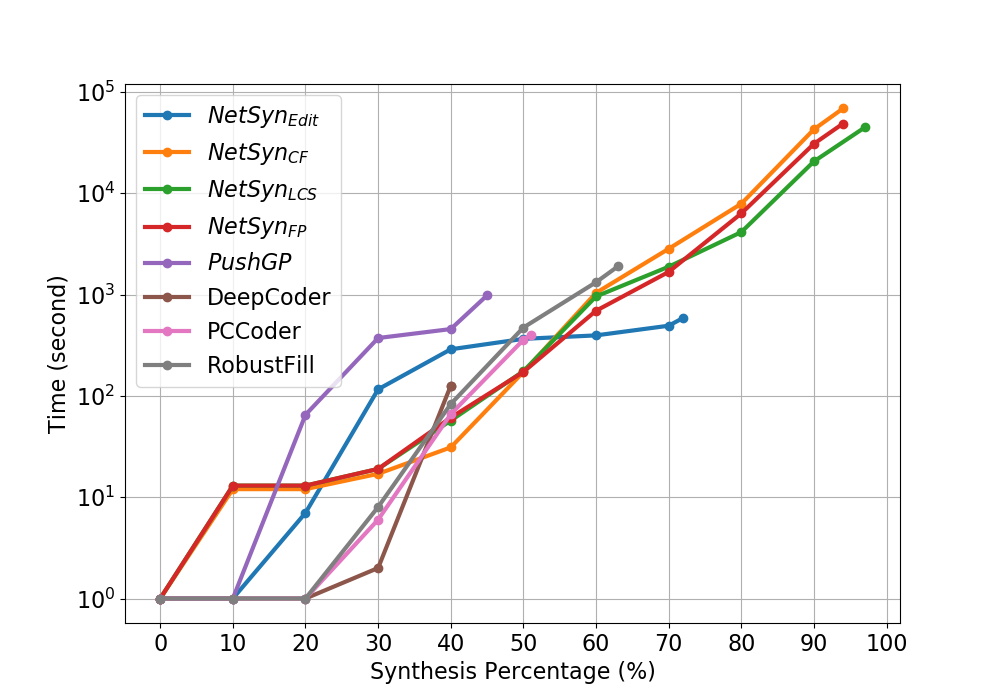}
    \vspace{-0.6cm}
    \caption{Program length = 5}
    \end{subfigure}
    ~
    \begin{subfigure}{0.3\textwidth}
        \includegraphics[width=\textwidth]{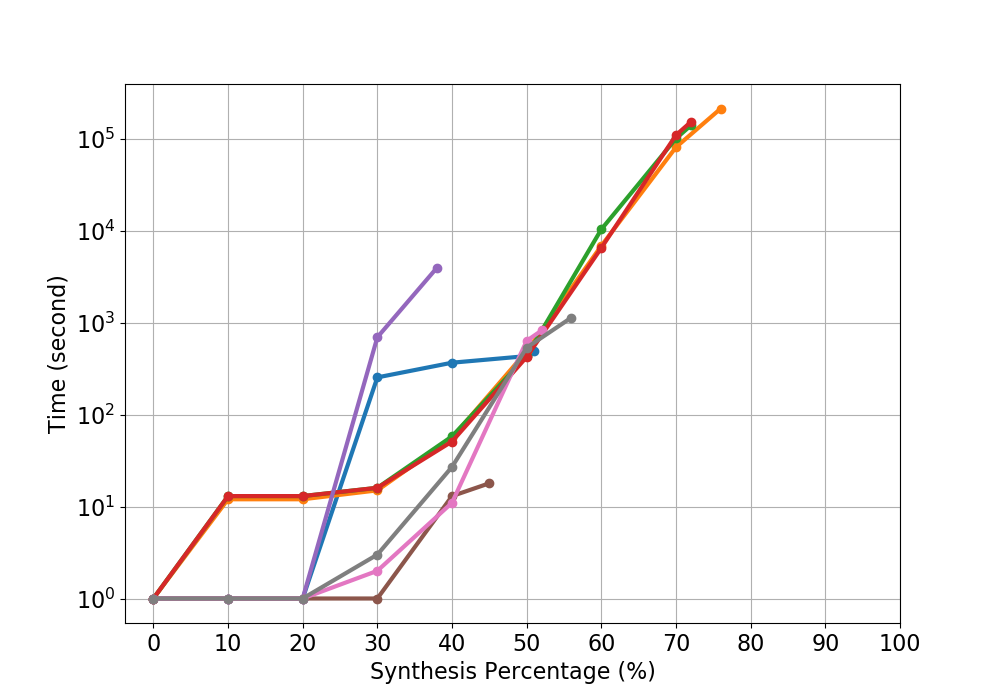}
     \vspace{-0.6cm}
     \caption{Program length = 7}
    \end{subfigure}
    ~
    \begin{subfigure}{0.3\textwidth}
        \includegraphics[width=\textwidth]{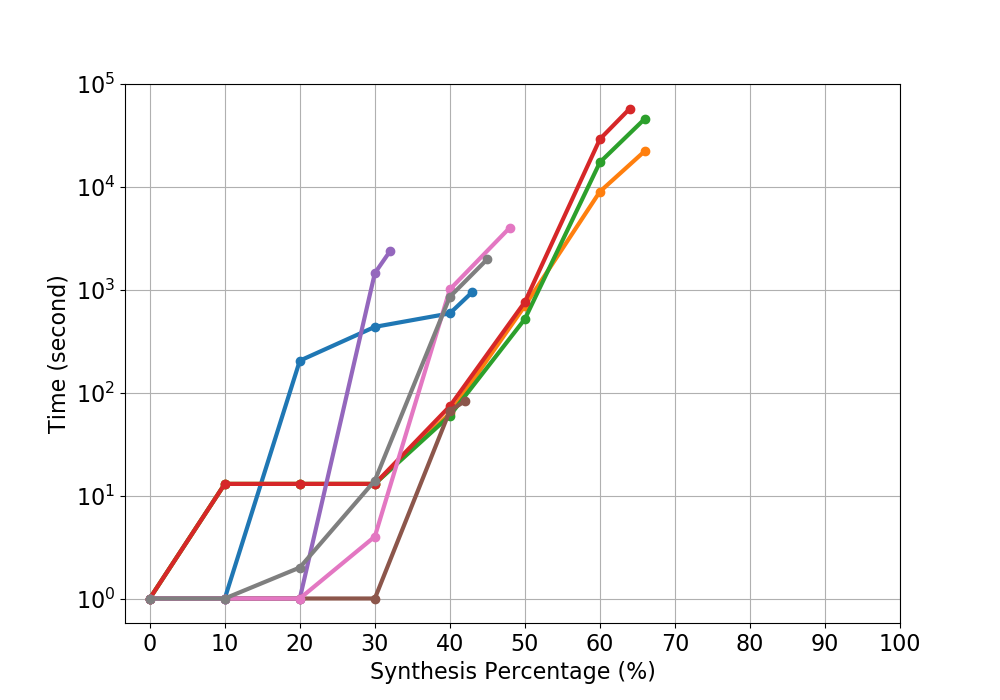}
     \vspace{-0.6cm}
     \caption{Program length = 10}
    \end{subfigure}        

    \caption{\scheme's synthesis ability with respect to different fitness functions and schemes. When limited by a maximum search space, \scheme\ synthesizes more programs than DeepCoder, PCCoder, RobustFill, and PushGP. Moreover for each program, \scheme\ synthesizes a higher percentage of runs than other approaches.}
    \label{fig:compare}
\end{figure*}

\section{Experimental Results}
\label{sec-results}

We implemented \scheme\ in Python with a TensorFlow backend~\cite{Abadi:2015:tensorflow}. We developed an interpreter for \scheme's DSL to evaluate the generated programs. We used 4,200,000 randomly generated unique example programs of length 5 to train the neural networks. We used 5 input-output examples for each program to generate the training data. To allow our model to predict equally well across all possible CF/LCS values, we generate these programs such that each of the 0-5 possible CF/LCS values for 5 length programs are equally represented in the dataset.  
To test \scheme, we randomly generated a total of 100 programs for each program length from 5 to 10. For each program length, 50 of the generated programs produce a singleton integer as the output; the rest produce a list of integers. We therefore refer to the first 50 programs as {\em singleton programs} and the rest as {\em list programs}. We collected $m=5$ input-output examples for each testing program.
When synthesizing a program using \scheme, we execute it $K = 10$ times and average the results to eliminate any noise.


\subsection{Demonstration of Synthesis Ability}
\label{sec-comparision}

We ran three variants of \scheme\ - $\mathit{\scheme_{CF}}$, $\mathit{\scheme_{LCS}}$, and $\mathit{\scheme_{FP}}$, each predicting $f^{\mathit{CF}}$, $f^{\mathit{LCS}}$, and $f^{\mathit{FP}}$ fitness functions, respectively. Each used $NS^{BFS}$ and FP-based mutation operation. 
We ran the publicly available best performing implementations of DeepCoder~\cite{deepcoder}, PCCoder~\cite{Zohar:2018:nips}, and RobustFill~\cite{robustfill}. We also implemented a genetic programming-based approach, PushGP~\cite{stackgp}.
For comparison, we also tested two other fitness functions: 
1) edit-distance between outputs ($f^{\mathit{Edit}}$), and 2) the oracle ($f^{\mathit{Oracle}}$). 
For every approach, we set the maximum search space size to 3,000,000 candidate programs. If an approach does not find the solution before reaching that threshold, we conclude the experiment marking it as \emph{``solution not found''}.

\begin{figure*}[htpb]
    \centering
    
    \begin{subfigure}{0.33\textwidth}
        \includegraphics[width=1.1\textwidth]{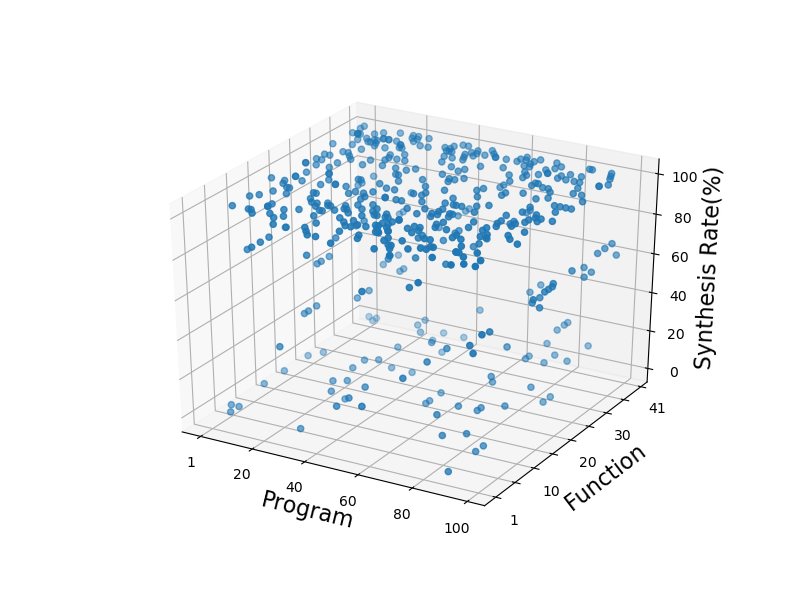}
    \vspace{-0.6cm}
    \caption{$\scheme_{CF}$}
    \end{subfigure}
    \begin{subfigure}{0.33\textwidth}
        \includegraphics[width=1.1\textwidth]{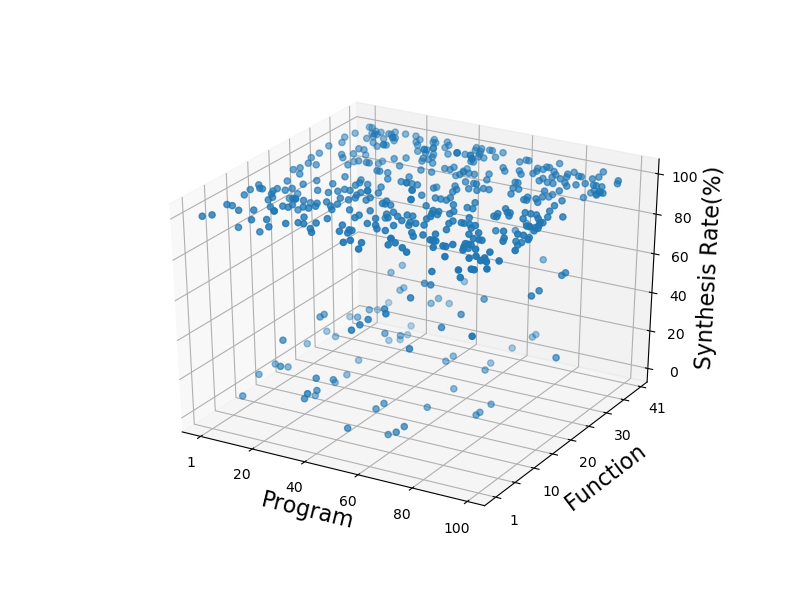}
     \vspace{-0.6cm}
     \caption{$\scheme_{LCS}$}
    \end{subfigure}
    \begin{subfigure}{0.33\textwidth}
        \includegraphics[width=1.1\textwidth]{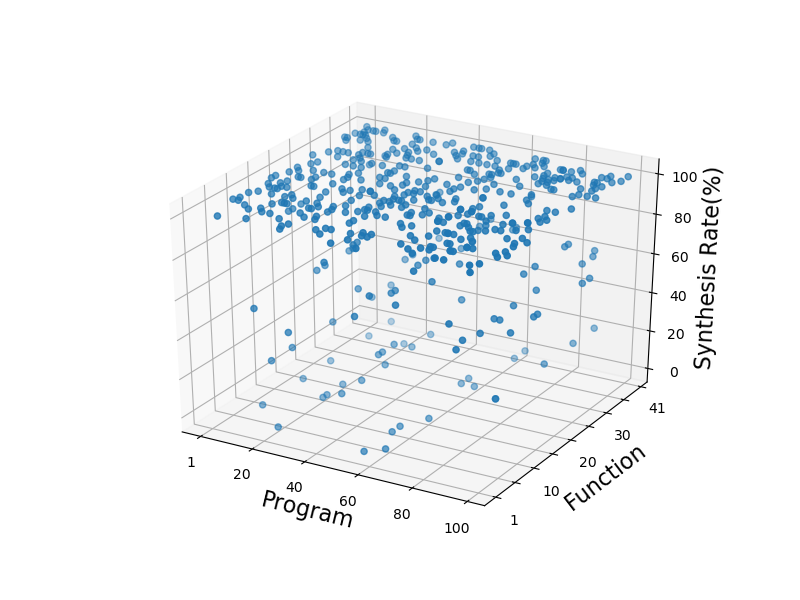}
     \vspace{-0.6cm}
     \caption{$\scheme_{FP}$}
    \end{subfigure}
    \caption{\scheme's synthesis ability with respect to fitness functions and DSL function types. Programs producing a single integer output are harder to synthesize in all three variants of \scheme.}
    \label{fig_total_found}
\end{figure*}

Figure~\ref{fig:compare}(a) - (c) show comparative results using the proposed metric: {\em search space} used. For each test program, we count the number of candidate programs searched before the experiment has concluded by either finding a correct program or exceeding the threshold. The number of candidate programs searched is expressed as a percentage of the maximum search space threshold, i.e., 3,000,000 and shown in y-axis. 
We sort the time taken to synthesize the programs. A position \emph{N} on the X-axis corresponds to the program synthesized in the \emph{Nth} longest percentile time of all the programs. Lines terminate at the point at which the approach fails to synthesize the corresponding program.
For all approaches, 
except for $f^{Edit}$-based \scheme\ and PushGP, up to 30\% of the programs can be synthesized by searching less than 2\% of the maximum search space. Search space use increases when an approach tries to synthesize more programs. In general, DeepCoder, PCCoder, and RobustFill search more candidate programs than $f^{\mathit{CF}}$, $f^{\mathit{LCS}}$ or $f^{\mathit{FP}}$-based \scheme. For example, for synthesizing programs of length 5, DeepCoder, PCCoder and RobustFill use 37\%, 33\%, and 47\% search space to synthesize 40\%, 50\%, and 60\% programs, respectively. In comparison, \scheme\ can synthesize upwards of 90\% programs by using less than 60\% search space. \scheme\ synthesizes programs at percentages ranging from 65\% (in case of $\mathit{\scheme_{FP}}$ for 10 length programs) to as high as 97\% (in case of $\mathit{\scheme_{LCS}}$ for 5 length programs).
In other words, \scheme\ is more efficient in generating and searching likely target programs. Even for length 10 programs, \scheme\ can generate 65\% of the programs using less than 45\% of the maximum search space. In contrast, DeepCoder, PCCoder, and RobustFill cannot synthesize more than 60\% of the programs even if they use the maximum search space. PushGP and edit distance-based approaches always use more search space than $f^{\mathit{CF}}$ or $f^{\mathit{LCS}}$. 

Figure~\ref{fig:compare}(d) - (f) show the distribution of synthesis rate (i.e., what percentage of $K=10$ runs synthesizes a particular program) in violin plots.
A violin plot shows interquartile range (i.e., middle 50\% range) as a vertical black bar with the median as a white dot. Moreover, wider section of the plot indicates more data points in that section.  
For 5 length programs, \scheme{} has a high synthesis rate (close to 100\%) for almost every program (as indicated by one wide section). On the other hand, DeepCoder, PCCoder, RobustFill, and PushGP have bimodal distributions as indicated by two wide sections.
At higher lengths, \scheme\ synthesizes around 65\% to 75\% programs and therefore, the distribution becomes bimodal with two wide sections. However, the section at the top is wider indicating that \scheme\ maintains high synthesis rate for the successful cases. DeepCoder, PCCoder, RobustFill, and PushGP have more unsuccessful cases than the successful ones. However, for the successful cases, these approaches also have high synthesis rates.

Figure~\ref{fig:compare}(g) - (i) show comparative results using synthesis time as the metric.  
In general, DeepCoder, PCCoder, RobustFill and \scheme\ can synthesize up to 20\% programs within a few seconds for all program lengths we tested. As expected, synthesis time increases as an approach attempts to synthesize more difficult programs. DeepCoder, PCCoder, and RobustFill usually find solutions faster than \scheme. 
It should be noted that the goal of \scheme\ is to synthesize a program with as few tries as possible. Therefore, the implementation of \scheme\ is not streamlined to take advantage of various parallelization and performance enhancement techniques such as GPUs, hardware accelerators, data parallel models etc.
The synthesis time tends to increase for longer length programs. 

\subsection{Characterization of \scheme}
\label{sec-effectiveness}

\begin{figure}[htb]
\begin{center}
\includegraphics[width=0.85\columnwidth]{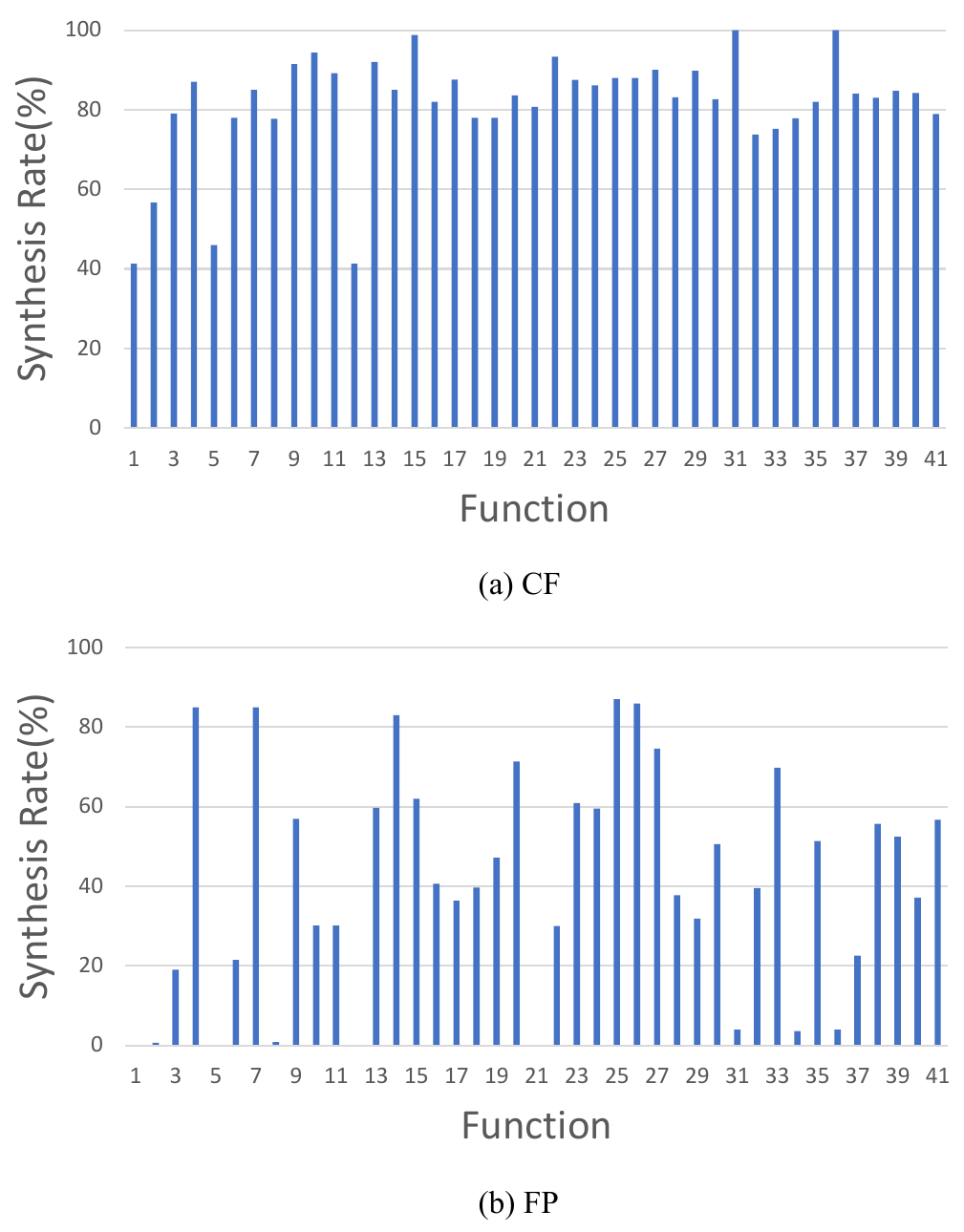}
\caption{Synthesis percentage across different functions. Functions 1 to 12 tend to have a lower synthesis rate because they produce a single integer output. Moreover, $f^{CF}$ has a higher synthesis rate.}
\label{fig_function}
\end{center}
\vspace{-0.6cm}
\end{figure}

Next, we characterize the effect of 
different components of \scheme.
We show the results in this section based on programs of length 5. However, we found our general observations to be true for longer length programs also.

\begin{table}[htpb]
   \centering
   \caption{Programs synthesized for different settings of \scheme. GA stands for genetic algorithm.}
\label{table:unique}
  \scalebox{0.7}{
  \begin{tabular}{l c  c  c}
  \hline
   Approach & Programs & Avg & Avg Syn.  \\
   & Synthesized & Generation & Rate (\%)  \\\hline
   $GA$ + $f^\mathit{CF}$   & 92 & 3273 &  74 \\
   $GA$ + $f^\mathit{CF}$ + $NS^{BFS}$  & 94 & 2953 & 77 \\
   $GA$ + $f^\mathit{CF}$ + $NS^{DFS}$ & 94 & 3026 & 76 \\
   $GA$ + $f^\mathit{CF}$ + $Mutation^{FP}$ & 93 & 2726 & 83 \\ 
   $GA$ + $f^\mathit{CF}$ + $NS^{BFS}$ + $Mutation^{FP}$ & 94 & 2275 & 85 \\\hline
  \end{tabular} }
\end{table}

\begin{figure*}[!ht]
    \centering
    
    \begin{subfigure}{0.33\textwidth}
        \includegraphics[width=1.1\textwidth]{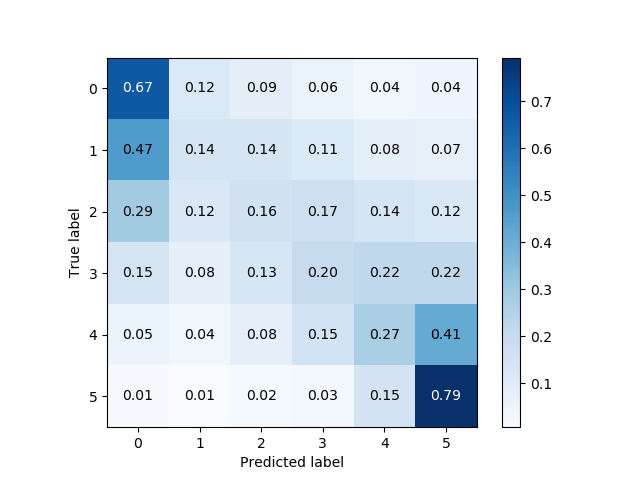}
    \vspace{-0.2cm}
    \caption{$f^{CF}$}
    \end{subfigure}
    \begin{subfigure}{0.33\textwidth}
        \includegraphics[width=1.1\textwidth]{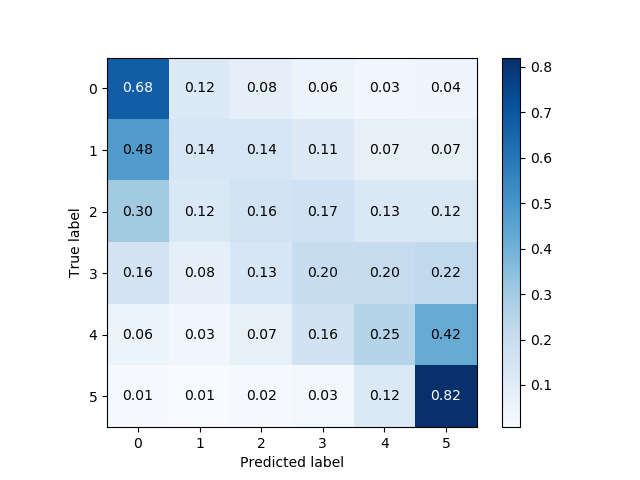}
     \vspace{-0.2cm}
     \caption{$f^{LCS}$}
    \end{subfigure}
    \begin{subfigure}{0.33\textwidth}
        \includegraphics[width=1.1\textwidth]{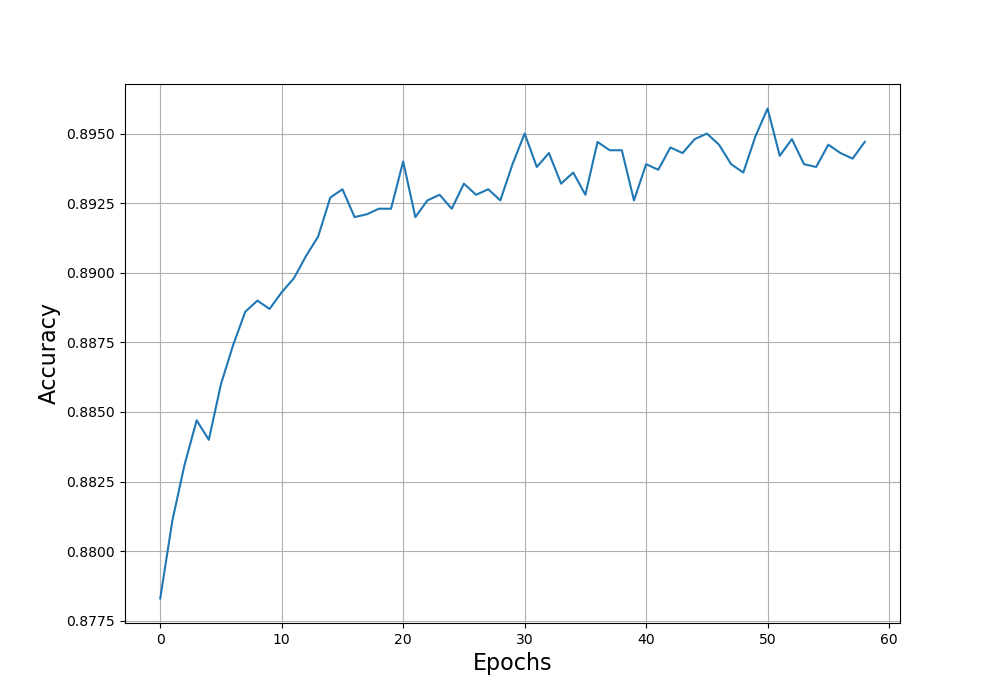}
     \vspace{-0.2cm}
     \caption{$f^{FP}$}
    \end{subfigure}
    \caption{Confusion matrix of (a) $f^{CF}$ (b) $f^{LCS}$ neural network fitness functions. (c) shows accuracy of  $f^{FP}$ over epochs. All graphs are based on the validation data. Overall, $f^{CF}$ and $f^{LCS}$ are capable identifying of close-enough solutions as well as mostly mistaken solutions. $f^{FP}$ reaches close to 90\% accuracy after 40 epochs.}
    \label{fig_total_found}
    \vspace{-0.2cm}
\end{figure*}

Table~\ref{table:unique} shows how many unique programs of $length=5$ (out of a total of 100 programs) that the different approaches were able to synthesize. It also shows the average generations and synthesis rate for each program. \scheme\ synthesized the most number of programs in the lowest number of generations and at the highest rate of synthesis when both the NS and improved mutation based on function probability ($Mutation^{FP}$) are used in addition to the the NN-FF. We note that BFS-based NS performs slightly better than DFS-based NS. Moreover, $Mutation^{FP}$ has some measurable impact on \scheme.
Figure~\ref{fig_total_found}(a) - (c) show the synthesis percentage for different programs and fitness functions. Program 1 to 50 are singleton programs and have lower synthesis percentage in all three fitness function choices. Particularly, the $f^{\mathit{FP}}$-based approach has a low synthesis percentage for singleton programs. Functions 1 to 12 produce singleton integer and tend to cause lower synthesis percentage for any program that contains them. 
This implies that singleton programs are relatively harder to synthesize.

To shed more light on this issue, Figure~\ref{fig_function} shows synthesis percentage across different functions. The synthesis percentage for a function is at least 40\% for the $f_{\mathit{CF}}$-based approach, whereas for the $f_{\mathit{FP}}$-based approach, four functions cannot be synthesized at all. 
Details of functions are in the appendix.


\subsection{Characterization of Neural Networks}
Figure~\ref{fig_total_found}(a), (b), and (c) show the prediction ability of our proposed
neural network fitness functions on validation data. Figure~\ref{fig_total_found}(a) \& (b) show the confusion matrix
for $f^{CF}$ and $f^{LCS}$ neural network fitness functions. The confusion matrix is a two dimensional matrix where $(i, j)$ entry indicates the probability of predicting the value $i$ when the actual value is $j$. Thus, each row of the matrix sums up to 1.0. We can see that when a candidate program is close to the solution (i.e., the fitness score is 4 or above), each of $f^{CF}$ and $f^{LCS}$-based model predicts a fitness score of 4 or higher with a probability of 0.7 or higher. In other words, the models are very accurate in identifying potentially close-enough solutions. Similar is the case when the candidate program is mostly mistaken (i.e., a fitness score is 1 or less). Thus, the neural networks are good at identifying both close-enough solutions and mostly wrong solutions. If a candidate program is some what correct (i.e., the candidate program has few correct functions but the rest of the functions are incorrect), it is difficult to identify them by the proposed models.

$f^{FP}$ model predicts probability of different functions given the IO examples. We assume a function probability to be correct if the function is in the target program and the neural network predicts its probability as 0.5 or higher. Figure~\ref{fig_total_found}(c) shows the accuracy of $f^{FP}$ model. With enough epochs, it reaches close to 90\% accuracy on the validation data set.

\subsubsection{Additional Models and Fitness Functions}
We tried several other models for neural networks and fitness functions. For example, instead of a classification problem, we treated fitness scores as a regression problem.  We found that the neural networks produced higher prediction error as the networks had a tendency to predict values close to the median of the values in the training set. With the higher prediction errors of the fitness function, the genetic algorithm performance degraded.

We also experimented with training a network to predict a correctness ordering among a set of genes.  We note that the ultimate goal of the fitness score is to provide an order among genes for the Roulette Wheel algorithm. Rather than getting this ordering indirectly via a fitness score for each gene, we attempted to have the neural network predict this ordering directly.  However, we were not able to train a network to predict this relative ordering whose accuracy was higher than the one for absolute fitness scores.  We believe that there are other potential implementations for this relative ordering and that it may be possible for it to be made to work in the future.

Additionally, we tried a two-tier fitness function.  The first tier was a neural network to predict whether a gene has a fitness score of 0 or not. In the event the fitness score was predicted to be non-zero, we used a second neural network to predict the actual non-zero value. This idea came from the intuition that since many genes have a fitness score of 0 (at least for initial generations), we can do a better job predicting those if we use a separate predictor for that purpose. Unfortunately, mispredictions in the first tier caused enough good genes to be eliminated that \scheme's synthesis rate was reduced.

Finally, we explored training a \emph{bigram} model (i.e., predicting pairs of functions appearing one after the other).  This approach is complicated by the fact that over 99\% of the $\mathit{41 \times 41}$ (i.e., number of DSL functions squared) bigram matrix are zeros. We tried a two-tiered neural network and principle component analysis to reduce the dimensionality of this matrix~\cite{li:2014:pca}. Our results using this bigram model in \scheme\ were similar to that of DeepCoder, with up to 90\% reduction in synthesis rate for singleton programs.


\section{Conclusion}
\label{sec-conc}

In this paper, we presented a genetic algorithm-based framework for program synthesis called \scheme. To the best of our knowledge, it is the first work that uses a neural network to automatically generate an genetic algorithm's fitness function in the context of machine programming. We proposed three neural network-based fitness functions. 
\scheme\ is also novel in that it uses neighborhood search to expedite the convergence process of a genetic algorithm. We compared our approach against several state-of-the art program synthesis systems -
DeepCoder~\cite{deepcoder}, PCCoder~\cite{Zohar:2018:nips}, RobustFill~\cite{robustfill}, and PushGP~\cite{stackgp}. 
\scheme\ synthesizes more programs than each of those prior approaches with fewer candidate program generations.
We believe that our proposed work could open up a new direction of research by automating fitness function generations for genetic algorithms by mapping the problem as a big data learning problem. This has the potential to improve any application of genetic algorithms.




\bibliographystyle{mlsys2020}
\bibliography{main}

\newpage
\appendix
\section{Appendix A: \scheme{}'s DSL}
\label{sec-appendix-dsl}
In this appendix, we provide more details about the list DSL that \scheme\ uses to generate programs.  Our list DSL has only two implicit data types, integer and list of integer.  A program in this DSL is a sequence of statements, each of which is a call to one of the 41 functions defined in the DSL.  There are no explicit variables, nor conditionals, nor explicit control flow operations in the DSL, although many of the functions in the DSL are high-level and contain implicit conditionals and control flow within them.  Each of the 41 functions in the DSL takes one or two arguments, each being of integer or list of integer type, and returns exactly one output, also of integer or list of integer type.  Given these rules, there are 10 possible function signatures.  However, only 5 of these signatures occur for the functions we chose to be part of the DSL.  The following sections are broken down by the function signature, wherein all the functions in the DSL having that signature are described.

Instead of named variables, each time a function call requires an argument of a particular type, our DSL's runtime searches backwards and finds the most recently executed function that returns an output of the required type and then uses that output as the current function's input.  Thus, for the first statement in the program, there will be no previous function's output from which to draw the arguments for the first function.  When there is no previous output of the correct type, then our DSL's runtime looks at the arguments to the program itself to provide those values.  Moreover, it is possible for the program's inputs to not provide a value of the requested type.  In such cases, the runtime provides a default value for missing inputs, 0 in the case of integer and an empty list in the case of list of integer.  For example, let us say that a program is given a list of integer as input and that the first three functions called in the program each consume and produce a list of integer.  Now, let us assume that the fourth function called takes an integer and a list of integer as input.  The list of integer input will use the list of integer output from the previous function call.  The DSL runtime will search backwards and find that none of the previous function calls produced integer output and that no integer input is present in the program's inputs either.  Thus, the runtime would provide the value 0 as the integer input to this fourth function call.  The final output of a program is the output of the last function called.

Thus, our language is defined in such a way that so long as the program consists only of calls to one of the 41 functions provided by the DSL, that these programs are valid by construction. 
Each of the 41 functions is guaranteed to finish in a finite time and there are no looping constructs in the DSL, and thus, programs in our DSL are guaranteed to finish.  This property allows our system to not have to monitor the programs that they execute to detect potentially infinite loops.  Moreover, so long as the implementations of those 41 functions are secure and have no potential for memory corruption then programs in our DSL are similarly guaranteed to be secure and not crash and thus we do not require any sand-boxing techniques.  When our system performs crossover between two candidate programs, any arbitrary cut points in both of the parent programs will result in a child program that is also valid by construction.  Thus, our system need not test that programs created via crossover or mutation are valid.

In the following sections, \emph{[]} is used to indicate the type list of integer whereas \emph{int} is used to indicate the integer type.  The type after the arrow is used to indicate the output type of the function.

\subsection{Functions with the Signature \emph{[]} $\rightarrow$ \emph{int}}
There are 9 functions in our DSL that take a list of integer as input and return an integer as output.
\subsubsection{HEAD (Function 6)}
This function returns the first item in the input list.  If the list is empty, a 0 is returned.
\subsubsection{LAST (Function 7)}
This function returns the last item in the input list.  If the list is empty, a 0 is returned.
\subsubsection{MINIMUM (Function 8)}
This function returns the smallest integer in the input list.  If the list is empty, a 0 is returned.
\subsubsection{MAXIMUM (Function 9)}
This function returns the largest integer in the input list.  If the list is empty, a 0 is returned.
\subsubsection{SUM (Function 11)}
This functions returns the sum of all the integers in the input list.  If the list is empty, a 0 is returned.
\subsubsection{COUNT (Function 2-5)}
This function returns the number of items in the list that satisfy the criteria specified by the additional lambda.  Each possible lambda is counted as a different function.  Thus, there are 4 COUNT functions having lambdas: >0, <0, odd, even.

\subsection{Functions with the Signature \emph{[]} $\rightarrow$ \emph{[]}}
There are 21 functions in our DSL that take a list of integer as input and produce a list of integer as output.
\subsubsection{REVERSE (Function 29)}
This function returns a list containing all the elements of the input list but in reverse order.
\subsubsection{SORT (Function 35)}
This function returns a list containing all the elements of the input list in sorted order.
\subsubsection{MAP (Function 19-28)}
This function applies a lambda to each element of the input list and creates the output list from the outputs of those lambdas.  Let $I_n$ be the nth element of the input list to MAP and let $O_n$ be the nth element of the output list from Map.  MAP produces an output list such that $O_n$=lambda($I_n$) for all n.  There are 10 MAP functions corresponding to the following lambdas: +1,-1,*2,*3,*4,/2,/3,/4,*(-1),\^{}2.
\subsubsection{FILTER (Function 14-17)}
This function returns a list containing only those elements in the input list satisfying the criteria specified by the additional lambda.  Ordering is maintained in the output list relative to the input list for those elements satisfying the criteria.  There are 4 FILTER functions having the lambdas: >0, <0, odd, even.
\subsubsection{SCANL1 (Function 30-34)}
Let $I_n$ be the nth element of the input list to SCANL1 and let $O_n$ be the nth element of the output list from SCANL1.  This function produces an output list as follows:

\[
\begin{cases} 
O_n=I_n~\&~n==0 \\
O_n=lambda(I_n,O_{n-1})~\&~n>0 
\end{cases}
\]

There are 5 SCANL1 functions corresponding to the following lambdas: +, -, *, min, max.

\subsection{Functions with the Signature \emph{int,{[]}} $\rightarrow$ \emph{[]}}
There are 4 functions in our DSL that take an integer and a list of integer as input and produce a list of integer as output.
\subsubsection{TAKE (Function 36)}
This function returns a list consisting of the first N items of the input list where N is the smaller of the integer argument to this function and the size of the input list.
\subsubsection{DROP (Function 13)}
This function returns a list in which the first N items of the input list are omitted, where N is the integer argument to this function.
\subsubsection{DELETE (Function 12)}
This function returns a list in which all the elements of the input list having value X are omitted where X is the integer argument to this function.
\subsubsection{INSERT (Function 18)}
This function returns a list where the value X is appended to the end of the input list, where X is the integer argument to this function.

\subsection{Functions with the Signature \emph{[],[]} $\rightarrow$ \emph{[]}}
There is only one function in our DSL that takes two lists of integers and returns another list of integers.
\subsubsection{ZIPWITH (Function 37-41)}
This function returns a list whose length is equal to the length of the smaller input list.  Let $O_n$ be the nth element of the output list from ZIPWITH.  Moreover, let $I^1_n$ and $I^2_n$ be the nth elements of the first and second input lists respectively.  This function creates the output list such that $O_n$=lambda($I^1_n$, $I^2_n$).  There are 5 ZIPWITH functions corresponding to the following lambdas: +, -, *, min, max.

\subsection{Functions with the Signature \emph{int,[]} $\rightarrow$ \emph{int}}
There are two functions in our DSL that take an integer and list of integer and return an integer.
\subsubsection{ACCESS (Function 1)}
This function returns the Nth element of the input list, where N is the integer argument to this function.  If N is less than 0 or greater than the length of the input list then 0 is returned.
\subsubsection{SEARCH (Function 10)}
This function return the position in the input list where the value X is first found, where X is the integer argument to this function.  If no such value is present in the list, then -1 is returned.

\section{Appendix B: System Details}
\label{sec-appendix-system}
\subsection{Hyper-parameters for the Models and Genetic Algorithm}
\begin{itemize}
\item{Evolutionary Algorithm:}
    \begin{itemize}
        \item{} Gene pool size: 100
        \item{} Number of reserve gene in each generation: 5
        \item{} Maximum number of generation: 30,000
        \item{} Crossover rate: 40\%
        \item{} Mutation rate: 30\%
    \end{itemize}

\end{itemize}

\section{Additional Results}
\begin{table*}[t]
    \caption{Comparison with DeepCoder and PCCoder in synthesizing different length programs. All experiments are done with the maximum search space set to 3,000,000 candidate programs.}
  \label{table:compare}
  \vskip 0.15in
  \begin{center}
  \begin{small}
  \begin{sc}
  \setlength\tabcolsep{1.1pt} 
  \begin{tabular}{clccccccccccc}
  \hline
  Program & \multirow{2}{*}{Method} & Synthesis & \multicolumn{10}{c}{Time Required to Synthesize (in seconds)} \\\cline{4-13}
  Length & & Percentage & 10\% & 20\% & 30\% & 40\% & 50\% & 60\% & 70\% & 80\% & 90\% & 100\%\\\hline
  \multirow{5}{*}{5}
& $\mathit{PushGP}$  & 45\%  & 1s  & 65s  & 372s  & 456s & - & - & - & - & - & - \\
& $\mathit{Edit}$  & 72\%  & 1s  & 7s  & 116s  & 288s  & 365s  & 395s  & 492s & - & - & - \\
& $\mathit{DeepCoder}$  & 40\%  & $<$1s  & $<$1s  & 2s  & 126s & - & - & - & - & - & - \\
& $\mathit{PCCoder}$  & 51\%  & 1s  & 1s  & 6s  & 66s  & 357s & - & - & - & - & - \\
& $\mathit{RobustFill}$  & 63\%  & 1s  & 1s  & 8s  & 83s  & 472s & 1321s & - & - & - & - \\
& $\mathit{\scheme_{FP}}$  & 94\%  & 13s  & 13s  & 19s  & 61s  & 172s  & 691s  & 1671s  & 6311s  & 30712s & - \\
& $\mathit{\scheme_{LCS}}$  & 97\%  & 13s  & 13s  & 19s  & 57s  & 175s  & 957s  & 1880s  & 4130s  & 20580s & - \\
& $\mathit{\scheme_{CF}}$  & 94\%  & 12s  & 12s  & 17s  & 31s  & 172s  & 1038s  & 2825s  & 7864s  & 42648s & - \\
& $\mathit{Oracle_{LCS|CF}}$  & 100\%  & $<$1s  & $<$1s  & $<$1s  & $<$1s  & $<$1s  & $<$1s  & 1s  & 1s  & 1s  & 1s \\
\hline

\multirow{5}{*}{7}
& $\mathit{PushGP}$  & 38\%  & 1s  & 1s  & 694s & - & - & - & - & - & - & - \\
& $\mathit{Edit}$  & 51\%  & 1s  & 1s  & 254s  & 367s  & 433s & - & - & - & - & - \\
& $\mathit{DeepCoder}$  & 45\%  & $<$1s  & $<$1s  & $<$1s  & 13s & - & - & - & - & - & - \\
& $\mathit{PCCoder}$  & 52\%  & 1s  & 1s  & 2s  & 11s  & 635s & - & - & - & - & - \\
& $\mathit{RobustFill}$  & 56\%  & 1s  & 1s  & 3s  & 27s  & 535s & - & - & - & - & - \\
& $\mathit{\scheme_{FP}}$  & 72\%  & 13s  & 13s  & 16s  & 51s  & 424s  & 6506s  & 109659s & - & - & - \\
& $\mathit{\scheme_{LCS}}$  & 72\%  & 13s  & 13s  & 16s  & 58s  & 433s  & 10363s  & 100728s & - & - & - \\
& $\mathit{\scheme_{CF}}$  & 76\%  & 12s  & 12s  & 15s  & 56s  & 489s  & 6862s  & 81037s & - & - & - \\
& $\mathit{Oracle_{LCS|CF}}$  & 100\%  & $<$1s  & $<$1s  & $<$1s  & $<$1s  & $<$1s  & $<$1s  & 1s  & 1s  & 1s  & 1s \\
\hline

\multirow{5}{*}{10}
& $\mathit{PushGP}$  & 32\%  & 1s  & 1s  & 1454s & - & - & - & - & - & - & - \\
& $\mathit{Edit}$  & 43\%  & 1s  & 205s  & 437s  & 591s & - & - & - & - & - & - \\
& $\mathit{DeepCoder}$  & 42\%  & $<$1s  & $<$1s  & $<$1s  & 67s & - & - & - & - & - & - \\
& $\mathit{PCCoder}$  & 48\%  & 1s  & 1s  & 4s  & 1011s & - & - & - & - & - & - \\
& $\mathit{RobustFill}$  & 45\%  & 1s  & 2s  & 14s  & 856s & - & - & - & - & - & - \\
& $\mathit{\scheme_{FP}}$  & 64\%  & 13s  & 13s  & 13s  & 74s  & 763s  & 29206s & - & - & - & - \\
& $\mathit{\scheme_{CF}}$  & 66\%  & 13s  & 13s  & 13s  & 63s  & 701s  & 9016s & - & - & - & - \\
& $\mathit{\scheme_{LCS}}$  & 66\%  & 13s  & 13s  & 13s  & 60s  & 521s  & 17384s & - & - & - & - \\
& $\mathit{Oracle_{LCS|CF}}$  & 100\%  & $<$1s  & $<$1s  & $<$1s  & $<$1s  & $<$1s  & $<$1s  & 1s  & 1s  & 1s  & 1s \\
\hline
  
  \end{tabular}
\end{sc}
\end{small}
\end{center}
\vskip -0.1in
\end{table*}

Table~\ref{table:compare} shows detailed numerical results using synthesis time as the metric. Columns 10\% to 100\% show the duration of time (in seconds) it takes to synthesize the corresponding percentage of programs. 

\begin{table*}[t]
\caption{Comparison with DeepCoder and PCCoder in terms of search space use. All experiments are done with the maximum search space set to 3,000,000 candidate programs.}
  \label{table:compare2}
  \begin{center}
\begin{small}
\begin{sc}
\setlength\tabcolsep{4pt} 
  \begin{tabular}{clcccccccccc}
  \hline
  Program & \multirow{2}{*}{Method} & \multicolumn{10}{c}{Search Space Used to Synthesize} \\\cline{3-12}
  Length & & 10\% & 20\% & 30\% & 40\% & 50\% & 60\% & 70\% & 80\% & 90\% & 100\% \\
  \hline
  \multirow{5}{*}{5}
& $\mathit{PushGP}$  & $<$1\%  & 9\%  & 60\%  & 67\% & - & - & - & - & - & - \\
& $\mathit{Edit}$  & $<$1\%  & $<$1\%  & 17\%  & 43\%  & 54\%  & 60\%  & 73\% & - & - & - \\
& $\mathit{DeepCoder}$  & $<$1\%  & 1\%  & 1\%  & 37\% & - & - & - & - & - & - \\
& $\mathit{PCCoder}$  & $<$1\%  & 1\%  & 1\%  & 7\%  & 33\% & - & - & - & - & - \\
& $\mathit{RobustFill}$  & $<$1\%  & 1\%  & 1\%  & 8\%  & 35\% & 47\% & - & - & - & - \\
& $\mathit{\scheme_{FP}}$  & $<$1\%  & $<$1\%  & $<$1\%  & $<$1\%  & 1\%  & 4\%  & 13\%  & 30\%  & 55\% & - \\
& $\mathit{\scheme_{LCS}}$  & $<$1\%  & $<$1\%  & $<$1\%  & $<$1\%  & 1\%  & 8\%  & 17\%  & 25\%  & 48\% & - \\
& $\mathit{\scheme_{CF}}$  & $<$1\%  & $<$1\%  & $<$1\%  & $<$1\%  & 1\%  & 10\%  & 22\%  & 40\%  & 58\% & - \\
& $\mathit{Oracle_{LCS|CF}}$  & $<$1\%  & $<$1\%  & $<$1\%  & $<$1\%  & $<$1\%  & $<$1\%  & $<$1\%  & $<$1\%  & $<$1\%  & $<$1\% \\
\hline

\multirow{5}{*}{7}
& $\mathit{PushGP}$  & $<$1\%  & $<$1\%  & 82\% & - & - & - & - & - & - & - \\
& $\mathit{Edit}$  & $<$1\%  & $<$1\%  & 34\%  & 48\%  & 69\% & - & - & - & - & - \\
& $\mathit{DeepCoder}$  & $<$1\%  & $<$1\%  & 1\%  & 3\% & - & - & - & - & - & - \\
& $\mathit{PCCoder}$  & $<$1\%  & $<$1\%  & 1\%  & 1\%  & 38\% & - & - & - & - & - \\
& $\mathit{RobustFill}$  & $<$1\%  & $<$1\%  & 1\%  & 2\%  & 35\% & - & - & - & - & - \\
& $\mathit{\scheme_{FP}}$  & $<$1\%  & $<$1\%  & $<$1\%  & $<$1\%  & 3\%  & 31\%  & 47\% & - & - & - \\
& $\mathit{\scheme_{LCS}}$  & $<$1\%  & $<$1\%  & $<$1\%  & $<$1\%  & 3\%  & 26\%  & 59\% & - & - & - \\
& $\mathit{\scheme_{CF}}$  & $<$1\%  & $<$1\%  & $<$1\%  & $<$1\%  & 4\%  & 31\%  & 56\% & - & - & - \\
& $\mathit{Oracle_{LCS|CF}}$  & $<$1\%  & $<$1\%  & $<$1\%  & $<$1\%  & $<$1\%  & $<$1\%  & $<$1\%  & $<$1\%  & $<$1\%  & $<$1\% \\
\hline

\multirow{5}{*}{10}
& $\mathit{PushGP}$  & $<$1\%  & $<$1\%  & 90\% & - & - & - & - & - & - & - \\
& $\mathit{Edit}$  & $<$1\%  & 20\%  & 43\%  & 56\% & - & - & - & - & - & - \\
& $\mathit{DeepCoder}$  & $<$1\%  & $<$1\%  & 1\%  & 9\% & - & - & - & - & - & - \\
& $\mathit{PCCoder}$  & $<$1\%  & $<$1\%  & 1\%  & 61\% & - & - & - & - & - & - \\
& $\mathit{RobustFill}$  & $<$1\%  & 1\%  & 4\%  & 58\% & - & - & - & - & - & - \\
& $\mathit{\scheme_{FP}}$  & $<$1\%  & $<$1\%  & $<$1\%  & $<$1\%  & 5\%  & 34\% & - & - & - & - \\
& $\mathit{\scheme_{CF}}$  & $<$1\%  & $<$1\%  & $<$1\%  & $<$1\%  & 4\%  & 36\% & - & - & - & - \\
& $\mathit{\scheme_{LCS}}$  & $<$1\%  & $<$1\%  & $<$1\%  & $<$1\%  & 4\%  & 40\% & - & - & - & - \\
& $\mathit{Oracle_{LCS|CF}}$  & $<$1\%  & $<$1\%  & $<$1\%  & $<$1\%  & $<$1\%  & $<$1\%  & $<$1\%  & $<$1\%  & $<$1\%  & $<$1\% \\
\hline

  \end{tabular} 
\end{sc}
\end{small}
\end{center}
\vskip -0.1in
\end{table*}

\end{document}